\documentclass[10pt,twocolumn,letterpaper]{article}

\usepackage{cvpr}  
\usepackage{graphicx}
\usepackage{adjustbox}
\usepackage{subcaption}
\usepackage{caption}
\usepackage{booktabs}
\usepackage{multirow}
\usepackage{makecell}
\usepackage{colortbl}
\usepackage[table]{xcolor}
\usepackage{geometry}
\usepackage{amssymb}
\usepackage{pifont}

\usepackage{etoolbox}
\makeatletter
\renewcommand{\@maketitle}{
   \newpage
   \null
   \iftoggle{cvprrebuttal}{\vspace*{-.3in}}{\vskip .375in}
   \begin{center}
      \iftoggle{cvprrebuttal}{{\large \bf \@title \par}}{{\Large \bf \@title \par}}
      \iftoggle{cvprrebuttal}{\vspace*{-20pt}}{\vspace*{8pt}}{
        \large
        \@author
        \par
      }
      \vskip .5em
      \vspace*{4pt}
   \end{center}
}
\makeatother

\newcommand{\cmark}{\textcolor{green!70!black}{\ding{51}}}
\newcommand{\xmark}{\textcolor{red}{\ding{55}}}

\definecolor{cvprblue}{rgb}{0.21,0.49,0.74}
\definecolor{lightorange}{RGB}{255, 230, 200}
\definecolor{middleorange}{RGB}{255, 205, 150}
\definecolor{orange}{RGB}{255, 180, 100}
\usepackage[pagebackref,breaklinks,colorlinks,allcolors=cvprblue]{hyperref}

\def\paperID{14689} 
\def\confName{CVPR}
\def\confYear{2026}

\title{
\adjustbox{valign=c,raise=0.3ex}{\includegraphics[height=3.5em]{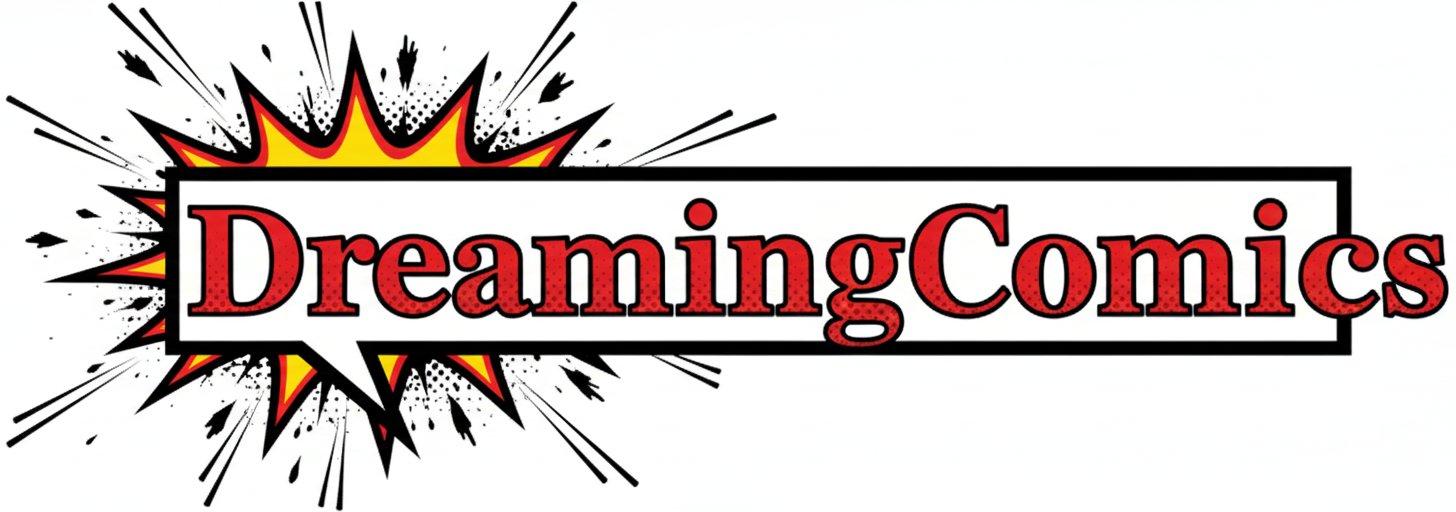}}:  A Story Visualization Pipeline via Subject and Layout Customized Generation using Video Models}

\author{Patrick Kwon, Chen Chen\\
Center for Research in Computer Vision, University of Central Florida, USA\\
{\tt\small yo564250@ucf.edu, chen.chen@crcv.ucf.edu}
}

\begin{document}

\twocolumn[{%
\renewcommand\twocolumn[1][]{#1}%
\maketitle

\begin{center}
    \centering
    \vspace{-8mm}
    \includegraphics[width=1.04\textwidth]{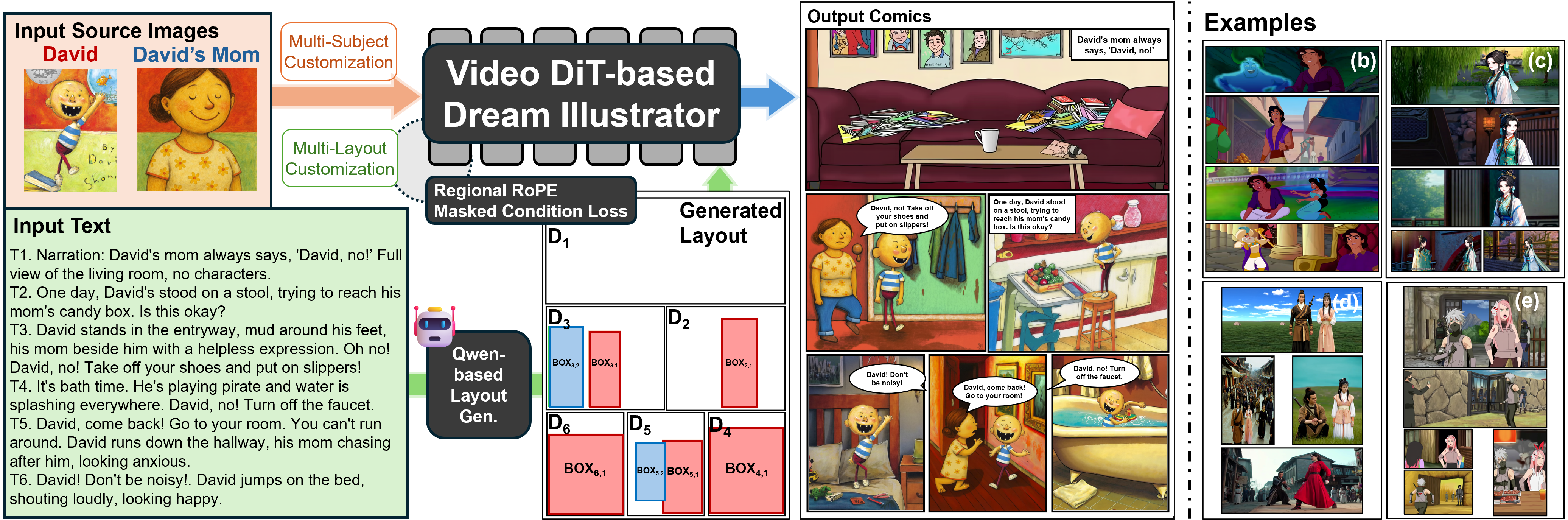}
    \vspace{-6mm}
    \captionof{figure}{(left) Overview of \textbf{DreamingComics}, a story visualization framework for multi subject and layout control. The dialogues are post-edited by humans. (right) Examples from DreamingComics, which is capable of generating layout-controlled stories with diverse art styles such as pencil illustrations, Disney-style animation, digital line-art, live-action drama, and cel animation.}
    \label{fig:title_figure}
\end{center}%
}]

\begin{abstract}

\vspace{-4mm}

Current story visualization methods tend to position subjects solely by text and face challenges in maintaining artistic consistency. To address these limitations, we introduce \textbf{DreamingComics}, a layout-aware story visualization framework. We build upon a pretrained video diffusion-transformer (DiT) model, leveraging its spatiotemporal priors to enhance identity and style consistency. For layout-based position control, we propose RegionalRoPE, a region-aware positional encoding scheme that re-indexes embeddings based on the target layout. Additionally, we introduce a masked condition loss to further constrain each subject's visual features to their designated region. To infer layouts from natural language scripts, we integrate an LLM-based layout generator trained to produce comic-style layouts, enabling flexible and controllable layout conditioning. We present a comprehensive evaluation of our approach, showing a 29.2\% increase in character consistency and 36.2\% increase in style similarity compared to previous methods, while displaying high spatial accuracy. Our project page is available at \url{https://yj7082126.github.io/dreamingcomics/}

\end{abstract}    
\section{Introduction}
\label{sec:intro}

\vspace{-2mm}
Story visualization~\cite{Xu2025MMStoryAgent, Cheng_2024_TheaterGen_arxiv, Zhou_2024_StoryDiffusion_NeurIPS, Mao_2024_StoryAdapter_arxiv}, the task of generating coherent visual sequences from textual narratives and character identities, has gained increasing attention with the advancement of generation models ~\cite{Rombach2022LDM, Podell2023SDXL, Esser_2024_SD3_ICML, BlackForestLabs2025FLUX} and controllable image customization methods \cite{Zhang2023ControlNet, Mou2023T2IAdapterLA, Ye2023IPAdapterTC, Wu_2025_UNO_ICCV, Mou_2025_DreamO_arxiv}. However, despite rapid improvements, existing approaches still lack the visual control necessary for storytelling tasks. 

Firstly, \textbf{controlling the position of multiple characters} is crucial for scene design, yet text prompts alone lack the pixel-level precision needed to specify spatial layout and character identity. Although recent image customization methods for diffusion transformer (DiT) models \cite{Peebles_2023_DiT_ICCV, Esser_2024_SD3_ICML, BlackForestLabs2025FLUX} can control who appears in the image \cite{Wu_2025_OmniGen2_arxiv, Wu_2025_UNO_ICCV, Mou_2025_DreamO_arxiv} or where objects are placed \cite{Zhang_2025_Eligen_arxiv, chen2024trainingfree}, they struggle to support both simultaneously, resulting in images with overlapping characters or incorrect appearances. Moreover, \textbf{preserving consistent artistic styles}, especially in cartoon and flat-shaded illustrations, remains difficult, as image generative models are typically biased toward photorealistic rendering, often overriding stylistic intent. In addition, the lack of paired datasets with both subject identity and position signals hinders the development of layout-aware customization. These challenges are most apparent in the comics domain, where consecutive image panels must preserve character identity and style while conforming to diverse layout structures unique to comics.

\begin{table}[!t]
    \centering
    \resizebox{\linewidth}{!}{
    \begin{tabular}{lcccc}
    \toprule
    \textbf{\makecell{Method}} & \textbf{\makecell{Layout Control}} & \textbf{\makecell{Multi-Subject \\Interaction}} & \textbf{\makecell{DiT \\architecture}} & \textbf{\makecell{LLM-based \\Layout}} \\
    \midrule
    Story-Adapter \cite{Mao_2024_StoryAdapter_arxiv}       & \xmark & \cmark & \xmark & \xmark \\
    StoryDiffusion \cite{Zhou_2024_StoryDiffusion_NeurIPS} & \xmark & \xmark & \xmark & \xmark \\
    TheaterGen \cite{Cheng_2024_TheaterGen_arxiv}          & \cmark & \xmark & \xmark & \cmark \\
    DiffSensei \cite{Wu_2025_DiffSensei_CVPR}              & \cmark & \cmark & \xmark & \xmark \\
    UNO \cite{Wu_2025_UNO_ICCV}                            & \xmark & \cmark & \cmark & \xmark \\
    DreamO \cite{Mou_2025_DreamO_arxiv}                    & \xmark & \cmark & \cmark & \xmark \\
    Eligen \cite{Zhang_2025_Eligen_arxiv}                  & \cmark & \xmark & \cmark & \xmark \\
    RealGeneral \cite{Lin_2025_RealGeneral_ICCV}           & \xmark & \xmark & \cmark & \xmark \\
    DRA-Ctrl \cite{Cao_2025_DraCtrl_arxiv}                 & \xmark & \xmark & \cmark & \xmark \\
    \rowcolor{gray!10}
    \textbf{DreamingComics (Ours)} & \cmark & \cmark & \cmark & \cmark \\
    \bottomrule
    \end{tabular}
    }
    \vspace{-3mm}
    \caption{Comparison of DreamingComics with prior works across various capabilities.}
    \label{tab:comparison}
    \vspace{-6mm}
\end{table}

In this paper, we present \textbf{DreamingComics}, a story visualization framework that simultaneously supports \textbf{multi-subject identity/style preservation and layout control}. We decompose the task into two modules: an LLM-based layout generator and a layout-aware customization model, \textit{Dream-Illustrator}. The layout generator produces comic-style layouts consisting of panels and character bounding boxes from a given storyline, while Dream-Illustrator renders consistent characters within those regions. This enables image customization to accurately position characters instead of solely relying on text, and enhance usability by reducing the burden on users of designing intricate layouts.


We build Dream‑Illustrator on a pretrained video DiT backbone \cite{Kong_2024_HunyuanVideo_arxiv, Zhang_2025_FramePack_arxiv}, which, unlike image‑specific generators, offers weaker perceptual quality but strong spatiotemporal priors that enhance visual consistency. Through FramePack \cite{Zhang_2025_FramePack_arxiv}, we repurpose a video model for image customization, enabling robust style generalization while keeping an image-level computation bottleneck. To control subject position by layout, we introduce \textbf{RegionalRoPE}, a regional 3D Rotary Position Embedding (RoPE) scheme that aligns each subject with its spatial region. To further improve layout fidelity, we employ a \textbf{masked condition loss} that penalizes attention beyond the assigned regions. Together, these mechanisms allow our model to generate identity-preserving images aligned with layouts. Lastly, we \textbf{construct an image-layout-paired dataset} for training Dream-Illustrator by annotating high-quality video samples. In parallel, we reformulate comic datasets into text-layout pairs to train our layout generator.


We evaluate our framework on benchmarks such as ViStoryBench~\cite{Zhuang_2025_VistoryBench_arxiv} and DreamBench~\cite{peng2024dreambench}. Extensive experiments demonstrate that DreamingComics can generate multi-subject customized images that closely align with target layouts while generalizing across a broad spectrum of artistic styles (Fig.~\ref{fig:title_figure} (b)-(e)).

In short, our main contributions are as follows.

\begin{itemize}
    \item DreamingComics, a layout-aware story visualization framework that supports multi-subject identity and style customization with spatial layout.
    \item An image customization model built on a pretrained video DiT model, leveraging spatiotemporal priors to improve style and identity consistency.
    \item Layout-aware conditioning strategies: RegionalRoPE, masked condition loss, and identity-layout-paired dataset generation that jointly enable accurate spatial control and reduce identity leakage.
    \item An LLM-guided layout generation module, fine-tuned on structured comic layout data, for prompt-based layout planning with minimal user input.
\end{itemize}

\section{Related Work}
\label{sec:related}

\begin{figure*}
  \centering
  \includegraphics[width=0.9\textwidth]{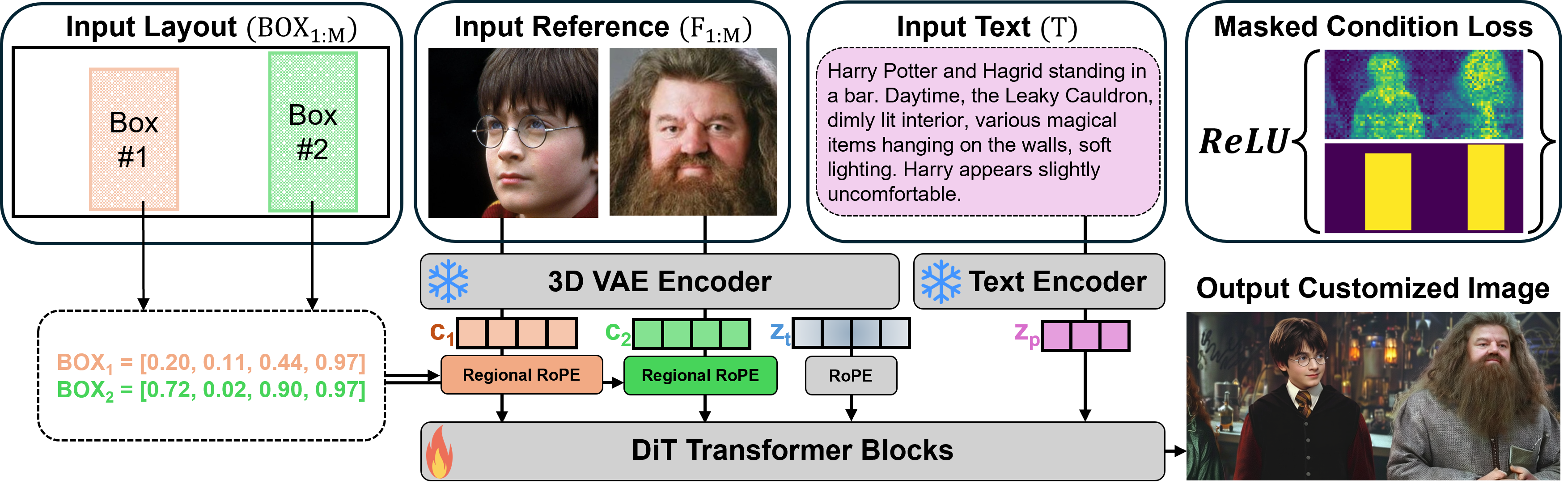}
  \vspace{-3mm}
  \caption{Overview of our image customization pipeline. The input reference images are encoded as token sequences $c_{1:n}$ along with the noise latent $z_t$ and the text latent $z_p$, passed to the stream of diffusion transformer blocks. We calculate a custom regional RoPE from the layout condition and apply it to the encoded references. During training, we calculate a Masked Condition Loss between the cross-attention map and the given layout condition, encouraging the model to position references within the layout.}
  \label{fig:main_arch}
  \vspace{-6mm}
\end{figure*}

\textbf{Image Customization.} Early methods for subject customization were based on finetuning the model for each specific subject \cite{Ruiz2022DreamBoothFT, Gal2022TextInversion, Hua_2023_DreamTuner_arxiv, Kumari2022MultiConceptCO}, achieving strong consistency but requiring training for each instance. Other tuning-free methods \cite{Ye2023IPAdapterTC, Dongxu_2023_BLIPDiffusion_NIPS, wei2023elite, zhang2024ssr} focused on injecting encoded results from an image encoder into a U-Net \cite{UNet} diffusion model, balancing efficiency with consistency. With the advent of new Diffusion Transformer (DiT) models \cite{Peebles_2023_DiT_ICCV, Esser_2024_SD3_ICML, BlackForestLabs2025FLUX}, many image customization works \cite{Wu_2025_UNO_ICCV, Tan_2024_OminiControl_arxiv, zhang2025easycontrol, Xiao_2024_OmniGen_arxiv, Mou_2025_DreamO_arxiv, Wu_2025_OmniGen2_arxiv, lhhuang2024iclora} directly reason over reference and noise latents in a unified attention space, improving subject fidelity without per-instance tuning. However, these approaches tend to rely on implicitly learned spatial cues, limiting spatial positioning.

There have been several studies on spatial position control for both training-based \cite{Li_2023_Gligen_CVPR, zhou2024migc, wang2024instancediffusion, Wang2024MSDiffusionMZ, Wang_2024_AttnMaskCtrl_AAAI} and training-free methods \cite{BarTal_2023_MultiDiffusion_ICML, ma2023directed, Cheng_2024_TheaterGen_arxiv} based on a pretrained U-Net diffusion model.  Recently, methods such as Regional Prompting \cite{chen2024trainingfree} and Eligen \cite{Zhang_2025_Eligen_arxiv} have proposed regional masking over the full-attention structure to control DiT models. However, explicit multi-subject spatial layout control remains underexplored, particularly within the DiT framework. Although TheaterGen \cite{Cheng_2024_TheaterGen_arxiv} supports multi-subject layout control, it generates each subject individually and merges them with ControlNet \cite{Zhang2023ControlNet}, omitting any interactions between subjects. 

Our work is in line with previous studies that employed video generative models for image-level tasks. For example, Frame2Frame \cite{Rotstein2024PathwaysOT} uses an image-to-video diffusion model \cite{Yang2024CogVideoX} for image editing by generating a sequence of frames, while FramePainter \cite{Zhang2025FramePainterEI} and Object-Mover \cite{Yu2025ObjectMoverGO} finetune a video model to perform object-level image editing tasks. Several works \cite{Chen2024UniRealUI, Lin_2025_RealGeneral_ICCV, Cao_2025_DraCtrl_arxiv} have also used video models for controllable image generation, demonstrating their capabilities to unify diverse image tasks. However, neither of them targeted multi-subject-driven generation aligned with a spatial layout. Our work uses novel strategies to incorporate layout control into a video model, utilizing multiple subject reference images with spatial layout.

\textbf{Story Visualization.} Recent advances in generative models and LLMs have resulted in significant progress in story visualization. Previous works on image-based story visualization \cite{Cheng2024AutoStudioCC, Liu_2024_StoryGen_CVPR, Pan2022SynthesizingCS, Wang2023AutoStoryGD, Yang_2024_SEEDStory_arxiv, Zhou_2024_StoryDiffusion_NeurIPS} focused on improving cross-frame alignment for images generated by a U-Net-based image generator. Comic generation \cite{Wu_2025_DiffSensei_CVPR, chen2024mangagenerationlayoutcontrollablediffusion} is also a notable subtask within story visualization. Although previous works such as DiffSensei \cite{Wu_2025_DiffSensei_CVPR} were restricted to monochrome manga aesthetics, our approach supports a wider range of artistic styles, from stylized cartoon art to color-heavy illustrations, allowing diverse and culturally agnostic storytelling.

Several recent works use large language models to reason about scene layouts from text~\cite{Lian2023LLMgroundedDE, Cheng_2024_TheaterGen_arxiv}. For instance, TheaterGen~\cite{Cheng_2024_TheaterGen_arxiv} and AutoStudio~\cite{Cheng2024AutoStudioCC} employ training-free LLM agents to predict image-level layouts. In contrast, our module is fine-tuned on curated comic data, enabling the generation of both panel-wise and page-level layouts—a granularity not addressed by prior work. While DiffSensei \cite{Wu_2025_DiffSensei_CVPR} also incorporates an MLLM for its inference, \textit{its usage is limited to an identity adapter and does not support layout generation.}

\vspace{-2mm}
\section{Methodology}

In Sec.~\ref{prelim}, we formulate our objectives and introduce our base model, HunyuanVideo-I2V \cite{Kong_2024_HunyuanVideo_arxiv}. In Sec.~\ref{layoutgen}, we introduce our LLM-based layout generator that outputs multi-panel layouts from text input. In Sec.~\ref{imagecustom}, we present Dream-Illustrator, an \textbf{image customization model based on a DiT video generator}, adapted for layout-aware next-frame prediction through a \textbf{regional RoPE scheme and masked condition loss}. In Sec.~\ref{datagen}, we report the dataset generation pipeline for both comic layout and image customization dataset.


\subsection{Problem Statement and Preliminaries} \label{prelim}
Our objective is to visualize stories through a comic-style narrative. Given a script $T = \{ T_1, T_2, ..., T_n \}$, the traditional goal of story visualization is to generate a sequence of corresponding images $\{ p_1, p_2, ..., p_n \}$ of the same size. Comics, however, are composed of \textit{image panels} that have different resolutions and locations within a single page \cite{Vivoli_2024_OneMPComic_arxiv}. Although several works use LLMs to predict layouts for a single image \cite{Lian2023LLMgroundedDE, Cheng_2024_TheaterGen_arxiv}, they do not predict layouts for multiple panels, making them insufficient for comic generation. In contrast, we adopt a comic-centric representation, where panel $i$, consisting of $n$ characters, is represented as a tuple

\vspace{-3mm}
\begin{equation}
(T_i, D_i, \{\text{BOX}_{i,1}, ..., \text{BOX}_{i, n} \}),
\end{equation}
\vspace{-4mm}

\noindent where $T_i$ is the input text, $D_i \in \mathbb{R}^4$ is the panel's bounding box, and $\text{BOX}_{i,j} \in \mathbb{R}^4$ are subject-level bounding boxes for each character $j$. This structure enables us to reason about both the positioning of panels within a page and the positioning of characters within panels, reflecting the unique narrative of comics.

To utilize its rich spatiotemporal priors for visual consistency, we base our image customization method on HunyuanVideo-I2V \cite{Kong_2024_HunyuanVideo_arxiv}, a video generation model composed of a causal 3D Variational Autoencoder \cite{Kingma2013AutoEncodingVB} (3DVAE, $\mathcal{E}$), a multimodal large language model (MLLM) encoder, and a diffusion transformer with a unified full attention mechanism across spatial and temporal tokens. Given an input video $x \in \mathbb{R}^{(4T+1) \times 3 \times 16 H \times 16 W}$, the VAE encodes it into a latent representation $y \in \mathbb{R}^{(T+1) \times 16 \times 2 H \times 2 W}$, which is then patchified into a sequence of visual tokens. The textual prompts $T_p$ are then encoded as tokens using the MLLM encoder and concatenated with the visual tokens to form a single sequence $z$, and passed into a transformer with 3D Rotary Position Embeddings (RoPE) \cite{Su_2024_RoFormer} to model positional relationships. For I2V generation, HunyuanVideo-I2V replaces the tokens of the first frame with the conditioning image tokens. The model is then optimized using a flow matching loss \cite{Lipman2022FlowMF}
\begin{equation}
\mathcal{L} = || v_\theta(\mathbf{y}_t, t, C_I, T_P) - (\epsilon - \mathbf{y})||^2,
\end{equation}
\noindent where $\epsilon \sim \mathcal{N}(0, I)$, $\mathbf{y}_t = (1-t)\mathbf{y} + t\epsilon$, and $v_\theta$ represent the parametric neural network.

\subsection{Layout Generation Pipeline} \label{layoutgen}

\begin{figure}[t]
    \resizebox{\columnwidth}{!}{
   \centering
    \includegraphics[width=0.49\textwidth]{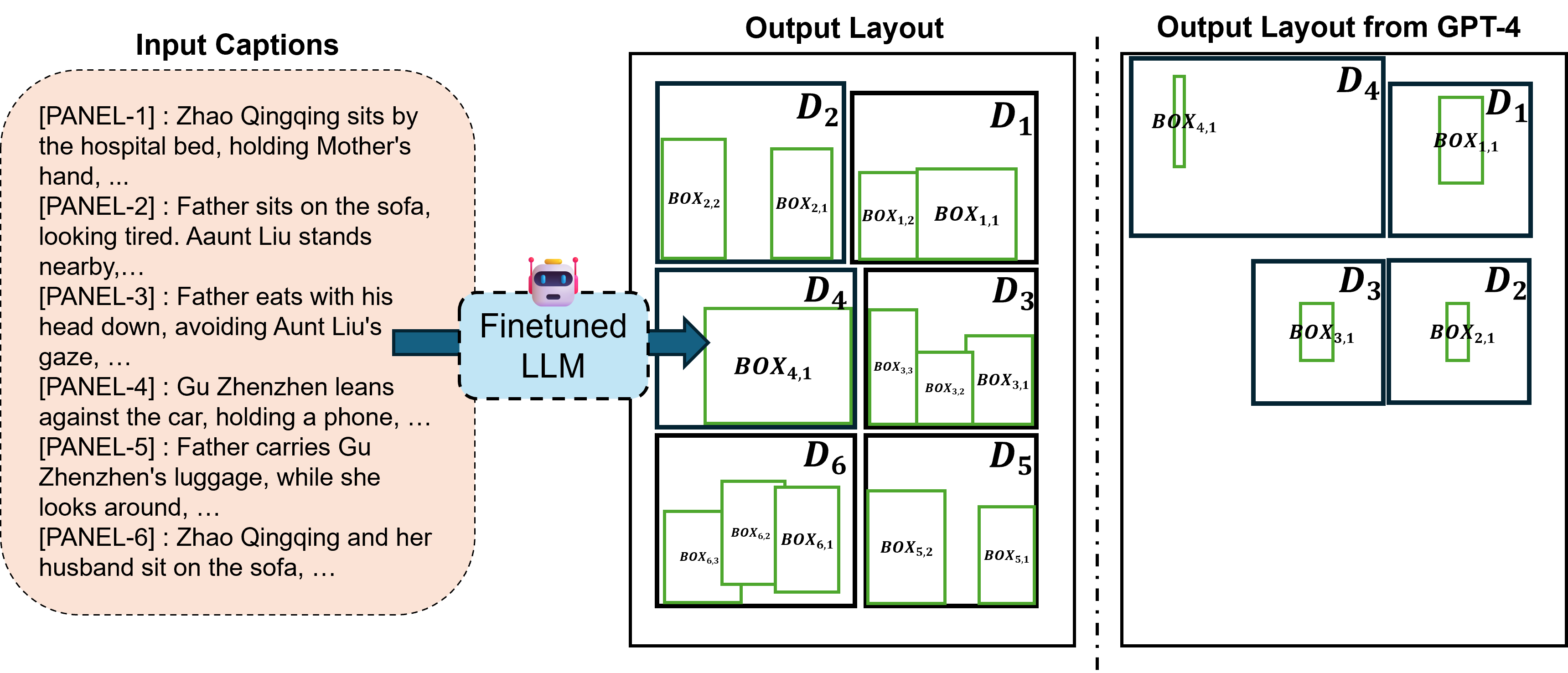}
}
  \vspace{-6mm}
  \caption{Given a list of textual descriptions for each panel, the finetuned LLM outputs a spatial layout for each panel and characters as a set of bounding boxes. Note that our layout, compared to the layout generated from the same prompt using GPT-4 \cite{Achiam2023GPT4TR}, occupies most of the panel region, correctly orders the panel (top-to-bottom, right-to-left), and draws plausible character boxes, constituting a ``good comic layout".}
  \vspace{-6mm}
  \label{fig:layoutgen}
\end{figure}

To generate page-level layouts from a given set of text descriptions, we fine-tune a large language model (LLM) \cite{yang2024qwen2technicalreport} using supervised fine-tuning (SFT) based on a newly crafted comics layout dataset. Given an input script $T$, the model is trained to predict a structured layout of panel regions and associated character positions, which are parsed as $(D_i, (\text{BOX}_{i,1}, ..., \text{BOX}_{i,n}))$. These representations are later used as layout conditions in the image customization stage, alongside their corresponding character images. A visualization of this layout generation process is shown in Fig.~\ref{fig:layoutgen}. By training an LLM with strong spatial reasoning over domain-specific data and making it understand the sparse visual cues within comics, we reduce the burden on users to manually specify intricate layouts while enabling the model to follow complex, prompt-driven visual instructions. 


\subsection{Dream-Illustrator} \label{imagecustom}

\textbf{Using FramePack as a next-frame predictor.} Since video generation models handle sequences of semantically consistent frames, they capture rich cross-frame context and complex spatiotemporal priors during training, which motivated us to use them for image customization; specifically, by using the reference image as the first frame to generate a subsequent target frame. Previous approaches in using video models for image tasks ~\cite{Rotstein2024PathwaysOT, Lin_2025_RealGeneral_ICCV, Cao_2025_DraCtrl_arxiv} required generating multiple frames for a single image output. At the same time, treating the problem as a simple two-frame video generation problem (generating the frame at $t=1$ conditioned on $t=0$) results in rigid ``copy-pasting" artifacts, with limited variation and poor prompt adherence. 

To address this, we build upon FramePack \cite{Zhang_2025_FramePack_arxiv}, a finetuned video DiT model that generates video frames in a progressive manner. Originally designed for efficient long-range video generation, FramePack compresses a sequence of $T$ input frames $(\mathbf{F} \in \mathbb{R}^{T \times h \times w \times c)}$ to sample $S$ output frames $(\mathbf{X} \in \mathbb{R}^{S \times h \times w \times c})$. Crucially, FramePack is capable of generating frames that are temporally distant from the input—for example, by producing an output at $t = 9$ from a reference at $t = 0$. We leverage this property to reformulate FramePack as an image customization module: given a target timestep $t'$ and a set of $N$ reference images $(\mathbf{F} = {F_1, F_2, ..., F_n})$  at $t = 0$, our goal is to generate a single target frame $\mathbf{X}$ that preserves subject identities while aligned with the input prompt. In practice, we encode each reference image $F_i$ into the latent space using VAE $(c_i = \mathcal{E}(F_i)$), and concatenate the patched latents with the latent noise $z_t$ and the latent text $z_p$. 

Compared to prior approaches~\cite{Chen2024UniRealUI, Lin_2025_RealGeneral_ICCV, Cao_2025_DraCtrl_arxiv}, our method leverages video model priors while generating a single frame, maintaining a relatively low computation bottleneck. As a result, we can infer a $1280 \times 720$ image within 17 seconds, over 3× faster than DRA-Ctrl ~\cite{Cao_2025_DraCtrl_arxiv}. Detailed comparisons of inference time and memory are provided in the \textbf{Supplementary}.


\begin{figure}[t]
    \centering
    \begin{subfigure}[t]{0.32\columnwidth}
        \centering
        \includegraphics[width=\textwidth]{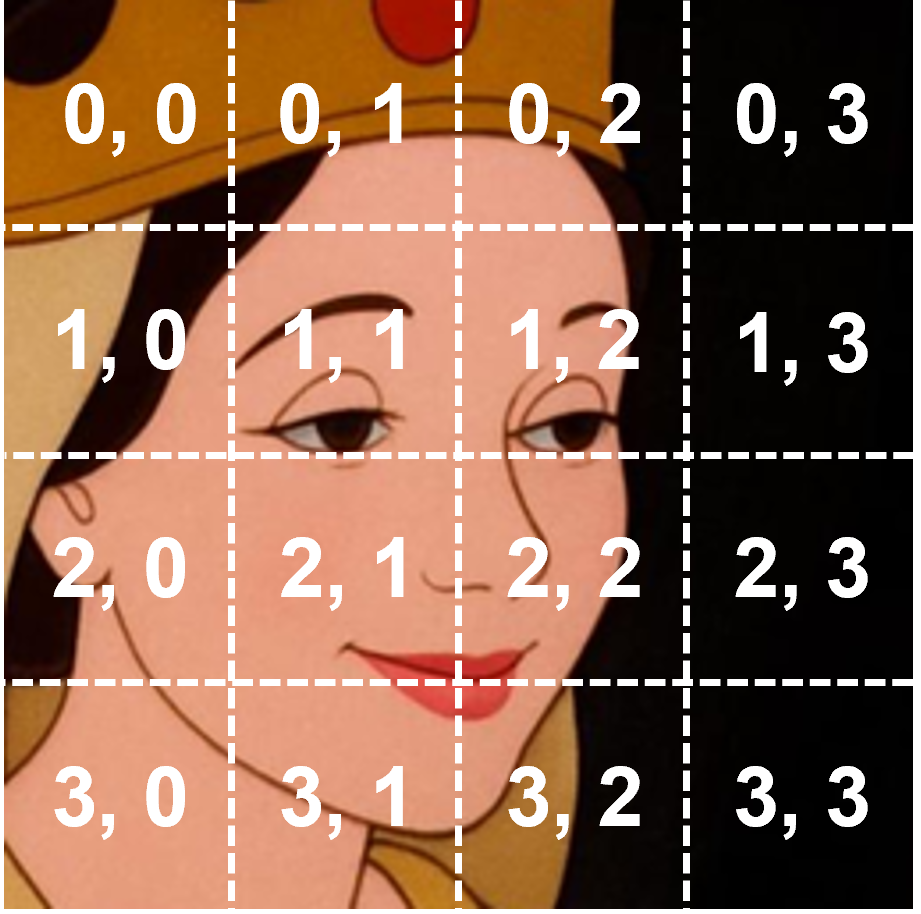}
        \subcaption{Reference Image with original RoPE}
    \end{subfigure}
    \begin{subfigure}[t]{0.56\columnwidth}
        \centering
        \includegraphics[width=\textwidth]{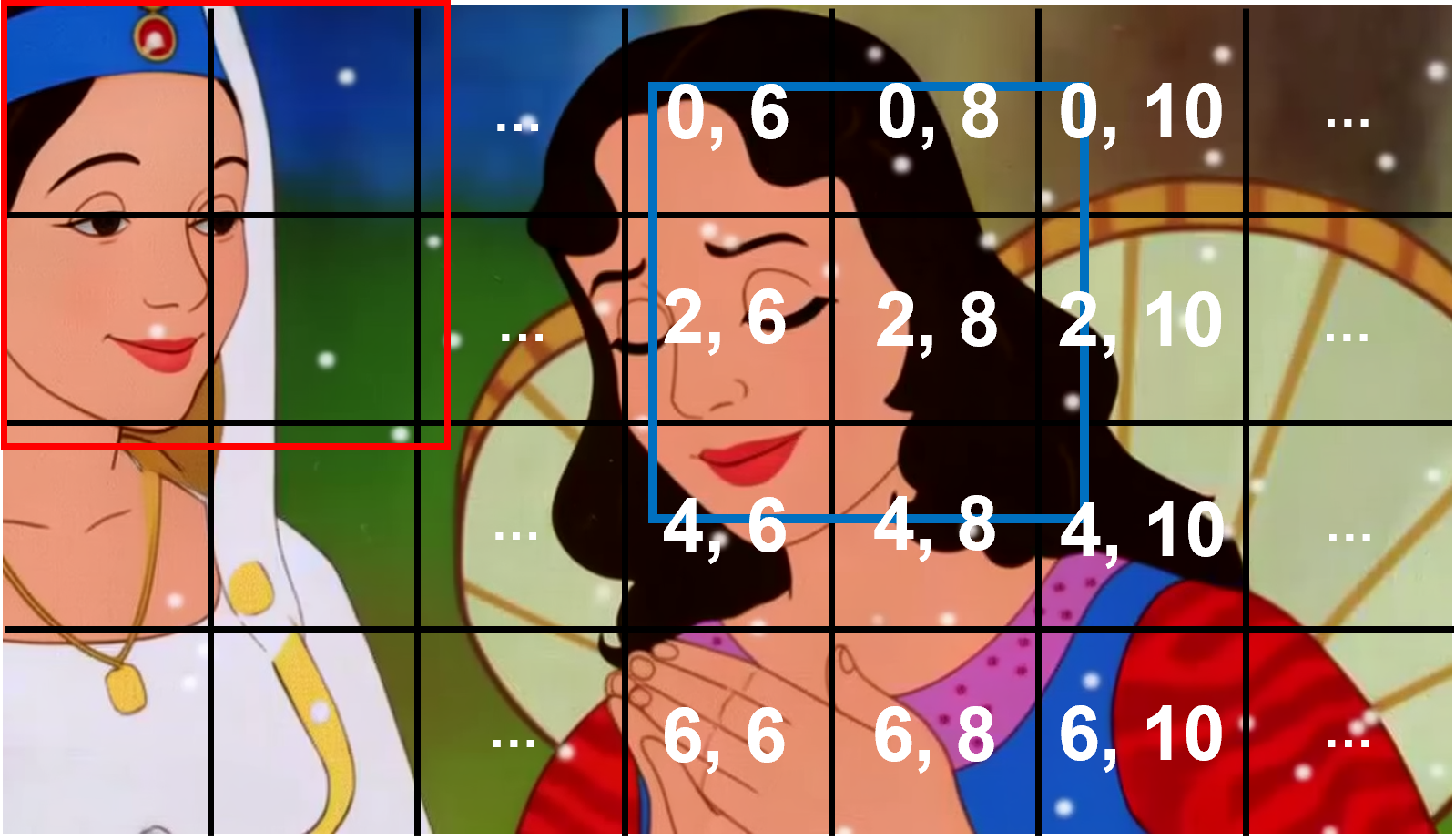}
        \subcaption{Generated Image}
    \end{subfigure}
    \begin{subfigure}[t]{0.32\columnwidth}
        \centering
        \includegraphics[width=\textwidth]{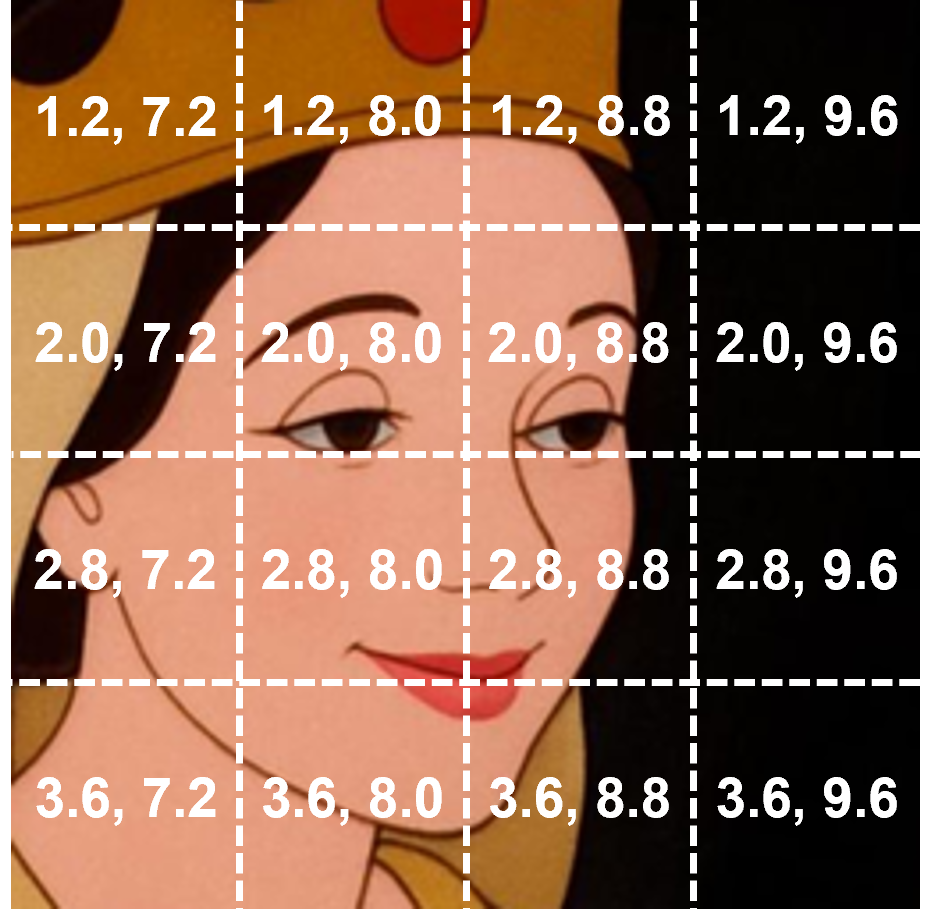}
        \subcaption{Reference Image with regional RoPE}
    \end{subfigure}
    \begin{subfigure}[t]{0.56\columnwidth}
        \centering
        \includegraphics[width=\textwidth]{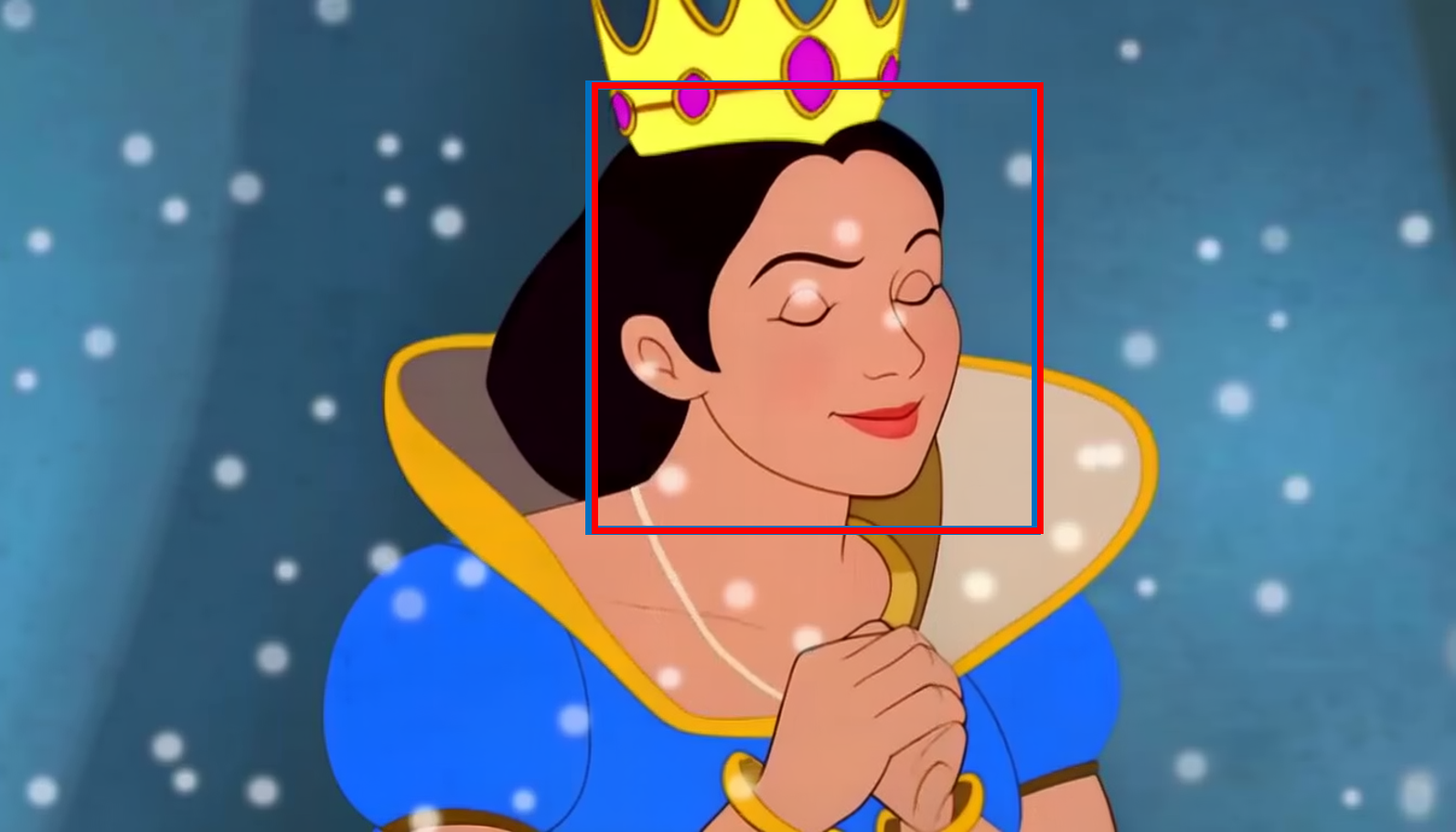}
        \subcaption{Generated Image}
    \end{subfigure}
  \vspace{-2mm}
  \caption{For illustration, we visualize the original positional indices for RoPE at (a) and the new indices at (c). The blue square indicates the intended layout region, and the red square indicates the actual generated region. The original RoPE restricts the reference content to the top-left corner, while ours can correctly position it according to the given layout.}
  \vspace{-6mm}
  \label{fig:regional_rope}
\end{figure}

\begin{figure*}[t]
    \centering
    \includegraphics[width=0.96\textwidth]{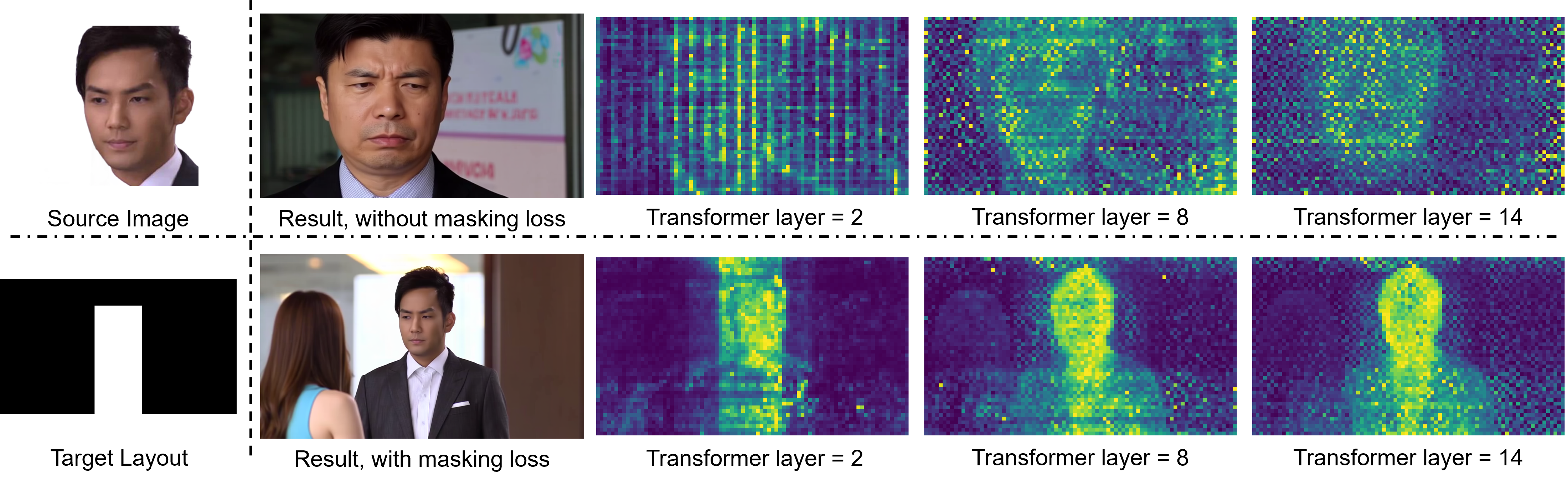}
    \vspace{-4mm}
      \captionof{figure}{Comparison between the $\text{CAMs}$ of our model trained without the masked condition loss (top) and with the masked condition loss (bottom). Using our new loss helps to position the attention around the target layout, which is evident in the third column (Transformer layer = 2), naturally inducing the model during training to position the character.
        }
    \vspace{-6mm}
  \label{fig:loss_figure}
\end{figure*}

\textbf{Regional RoPE for layout control.} The default RoPE assigns the same starting coordinates $(0, 0)$ to all reference frames, which makes the model perceive all subjects as if they are originating from the same region, leading to spatial entanglement and identity collapse (Fig.~\ref{fig:regional_rope}). To enable precise layout-driven spatial control, we introduce RegionalRoPE, a deterministic mapping of each reference’s RoPE indices to its target layout box. Unlike previous works that modify RoPE for spatial decorrelation \cite{Wu_2025_UNO_ICCV, Tan_2024_OminiControl_arxiv, Tan2025OminiControl2EC}, we use RegionalRoPE for explicit spatial grounding.

As mentioned in Sec.~\ref{prelim}, we have a region $\text{BOX}_i = [w_\text{start}, h_\text{start}, w_\text{end}, h_\text{end}]$ for each reference latent $c_i \in \mathbb{R}^{h_i \times w_i \times d}$. For the region size $(W_\text{box}, H_\text{box})$, to preserve the latent aspect ratio of the reference while fitting within the region, we compute a scaling factor $s = \min\!\left(\frac{W_\text{box}}{w_i}, \frac{H_\text{box}}{h_i}\right)$ and define the adjusted RoPE grid size as $(W', H') = (s\,w_i,\, s\,h_i)$. The RoPE grid is then positioned within the layout region with ranges
\begin{align}
w'_\text{start} &= w_\text{start} + \frac{W_\text{box} - W'}{2}, \; w'_\text{end} = w'_\text{start} + W' \\
h'_\text{start} &= h_\text{start} + a\, (H_\text{box} - H'), \; h'_\text{end} = h'_\text{start} + H'
\end{align}
\noindent where $a \in [0,1]$ controls the vertical alignment ($a{=}0$ for top-aligned, $a{=}0.5$ for center-aligned). The indices for each latent pixel $(i,j)$ are finally mapped as

\vspace{-2mm}
\begin{equation}
(t', i', j') = \Big(0,\;
w'_\text{start} + \frac{W'}{w_i} i,\;
h'_\text{start} + \frac{H'}{h_i} j \Big)
\end{equation}
\vspace{-4mm}

Each reference latent is encoded independently with its own coordinates, and the resulting sequences are concatenated into the input stream along with the target tokens. Previous works like UNO~\cite{Wu_2025_UNO_ICCV} and Ominicontrol~\cite{Tan_2024_OminiControl_arxiv, Tan2025OminiControl2EC} modify RoPE to decorrelate spatial cues and reduce reference copying, but they do not enforce explicit spatial alignment. In contrast, we assign RoPE offsets for each reference based on its region, enabling precise location mapping. Unlike other works ~\cite{Lin_2025_RealGeneral_ICCV, Cao_2025_DraCtrl_arxiv} which resize inputs to fixed frame sizes, we operate on native-resolution cropped latents, preserving subject fidelity and improving efficiency.




\textbf{Masked condition loss.} While RegionalRoPE aligns latents without extra training, its naive application can still lead to identity distortion and copy-paste artifacts. To address this, we introduce a masked condition loss that supervises each subject's spatial attention. During diffusion, we extract the cross-attention map $(\text{CAM})$ between the reference and the generated result as 

\vspace{-4mm}
\begin{equation}
\text{CAM}_{c_i, t, \text{block}_j} = \frac{Q_{c_i,  t, \text{block}_j} K_{t, \text{block}_j}^T}{\sqrt{d}}
\end{equation}
\vspace{-3mm}

\noindent where $Q_{c_i,  t, \text{block}_i}$ are the tokens of the $i$-th reference image and $K_{t, \text{block}_j}$ are the tokens of the noisy latent at timestep $t$, DiT layer $j$. Averaging these across the timestep and normalizing to $[0, 1]$ yields $\text{CAM}_{c_i, \text{block}_j}$, visualizing each subject's attention region.

\begin{table*}[ht!]
    \centering
    \resizebox{0.95\textwidth}{!}{
    \begin{tabular}{@{}lccccccccc@{}}
        \toprule
        \textbf{Method} & \textbf{Base Model} & \multicolumn{2}{c}{\textbf{CIDS (Char) ↑}} & \multicolumn{2}{c}{\textbf{CSD (Style) ↑}} & \textbf{OCCM ↑} & \textbf{Layout $\uparrow$} & \textbf{Inception ↑} & \textbf{Aesthetics ↑} \\
         & & Cross & Self & Cross & Self & Score & Precision & Score & Score \\
        \midrule
        
        Copy-Paste Baseline & - & 94.1 & 98.7 & 72.8 & 71.5 & 99.7 & - & 6.72 & 4.48 \\
        
        \addlinespace[0.5ex]
        Story-Adapter \cite{Mao_2024_StoryAdapter_arxiv} & SDXL \cite{Podell2023SDXL} & 35.3 & 57.4 & 29.6 & \underbar{63.0} & 78.8 & - & \textbf{9.79} & \underbar{5.55} \\
        StoryDiffusion \cite{Zhou_2024_StoryDiffusion_NeurIPS} & SDXL \cite{Podell2023SDXL} & 30.1 & 46.9 & 31.8 & 52.2 & 67.7 & - & 7.36 & 4.96 \\
        DiffSensei \cite{Wu_2025_DiffSensei_CVPR} & SDXL \cite{Podell2023SDXL} & 47.5 & 62.5 & 31.5 & \textbf{65.2} & \underbar{85.9} & \underbar{42.0} & 5.53 & 4.47 \\
        Eligen \cite{Zhang_2025_Eligen_arxiv} & FLUX \cite{BlackForestLabs2025FLUX} & 35.9 & 56.0 & 29.7 & 61.7 & 78.5 & 39.7 & 8.49 & \textbf{5.81} \\
        UNO \cite{Wu_2025_UNO_ICCV} & FLUX \cite{BlackForestLabs2025FLUX} & 46.2 & 61.1 & \underbar{39.3} & 59.3 & 83.8 & - & 8.42 & 5.20 \\
        DreamO \cite{Mou_2025_DreamO_arxiv} & FLUX \cite{BlackForestLabs2025FLUX} & \underbar{51.6} & \underbar{63.1} & 38.3 & 58.8 & 85.7 & - & 7.86 & 5.51 \\
        
        \addlinespace[0.5ex]
        Vlogger \cite{Zhuang_2024_Vlogger_CVPR} & Vlogger \cite{Zhuang_2024_Vlogger_CVPR} & 33.9 & 54.2 & 28.3 & 47.2 & 76.6 & - & 7.46 & 4.25 \\
        RealGeneral \cite{Lin_2025_RealGeneral_ICCV} & CogVideoX \cite{Yang2024CogVideoX} & 37.5 & 54.2 & 38.6 & 52.7 & 78.1 & - & 7.82 & 5.08 \\
        DRA-Ctrl \cite{Cao_2025_DraCtrl_arxiv} & HunyuanVid \cite{Kong_2024_HunyuanVideo_arxiv} & 36.2 & 50.2 & 39.0 & 58.0 & 74.9 & - & \underbar{8.77} & 5.03 \\
        Ours & FramePack \cite{Zhang_2025_FramePack_arxiv} & \textbf{66.6} & \textbf{68.1} & \textbf{53.6} & 58.0 & \textbf{86.7} & \textbf{61.6} & 6.81 & 4.54 \\
        \bottomrule
    \end{tabular}}
    \vspace{-2mm}
    \captionof{table}{Quantitative evaluation results on ViStoryBench. The \textbf{bold} value is the best, and the \underbar{underlined} value is the second best.}
    \label{tab:vistorybench}
    \vspace{-6mm}
\end{table*}

To ensure that these spatial attentions respect their layout boundaries, we introduce a ReLU-based loss that operates on each $\text{CAM}$ and the subject-wise binary spatial mask $\text{MASK}_i$. Here, $\text{MASK}_i$ is a binary mask from the bounding box $\text{BOX}_i$, with in-layout pixels represented as 1 and others as 0. The masked condition loss, with respect to the DiT layer 2, is then defined as

\vspace{-4mm}
\begin{equation}
\mathcal{L}_{\text{mask}} = \frac{1}{n_c}\sum_{i=1}^{n_c} \text{ReLU} ( \text{CAM}_{c_i, \text{block}_2} - \text{MASK}_i )
\end{equation}
\vspace{-4mm}

\noindent where $n_c$ refers to the number of conditions. Here, the ReLU penalizes only the attention leakage beyond the defined layout region while preventing unnecessary suppression of valid in-region focus. The final loss combines this term with the diffusion objective

\vspace{-2mm}
\begin{equation}
\mathcal{L} = \mathcal{L}_\text{diff} + \lambda_{\text{mask}} \mathcal{L}_\text{mask}
\end{equation}
\vspace{-4mm}

\noindent This constraint guides the model to respect spatial boundaries and maintain subject-specific attention, mitigating identity bleeding and visual artifacts. The effects of our new training loss are illustrated in Fig.~\ref{fig:loss_figure}.

Additionally, motivated by previous works \cite{Lin_2025_RealGeneral_ICCV, Cao_2025_DraCtrl_arxiv} for multiple subjects, we apply attention masking such that each reference latent should not be visible to one another to avoid information leakage. More details on this feature are provided in the \textbf{Supplementary}.

\subsection{Dataset Generation Pipeline} \label{datagen}

\textbf{Comics layout dataset.} To train our layout generator, we compile and annotate data from three comic datasets: COMICS \cite{Iyyer2016Comics}, Manga109 \cite{Fujimoto2016Manga109DA}, and PopManga \cite{Sachdeva2024TheMW}, capturing varied styles and eras in comics. For datasets lacking annotations, we apply the MagiV2 panel detector \cite{Sachdeva_2024_Magiv2_ACCV} to extract panel and character bounding boxes. All data are standardized, with inconsistent samples filtered out. We generate panel and page-wise descriptions using Qwen2.5-VL \cite{yang2024qwen2technicalreport}. 


\textbf{Paired subject dataset.} Public datasets rarely provide paired samples with references and layout conditions, which are critical for training Dream-Illustrator. To address this, we design a novel data framework that samples image pairs from curated video sources. Specifically, we build on OpenS2V-Nexus \cite{Yuan_2025_OpenS2VNexus_arxiv}, a large-scale dataset with structured annotations. We select videos with at least one consistently framed human subject. Using the segmentation map for the first frame, we extract its human bounding boxes as target layout conditions. The source frame is then chosen from a distant timestamp, ensuring subject continuity through facial presence. Samples are retained only if both the target (TopIQ) and the source face (TopIQ-Face) \cite{Chen_2024_TopIQ_TIP} scores exceed the quality thresholds. To cover diverse artistic styles, we apply a similar procedure to Anime-Shooter \cite{Qiu_2025_AnimeShooter_arxiv}, a reference-guided multi-shot animation dataset. Additionally, we processed the high-quality subset of Subject200K \cite{Tan_2024_OminiControl_arxiv} following DreamO \cite{Mou_2025_DreamO_arxiv}, using LISA to predict subject masks for the construction of paired data. All datasets are publicly released with their usage restricted to the research community, and we have fully adhered to their usage guidelines. In total, we compiled 55K single-subject and 20K multi-subject paired samples, with further details and ethical considerations provided in the \textbf{Supplementary}.



\section{Experiments}





\subsection{Implementation Details}

\begin{figure*}
    \centering
    \includegraphics[width=0.96\textwidth]{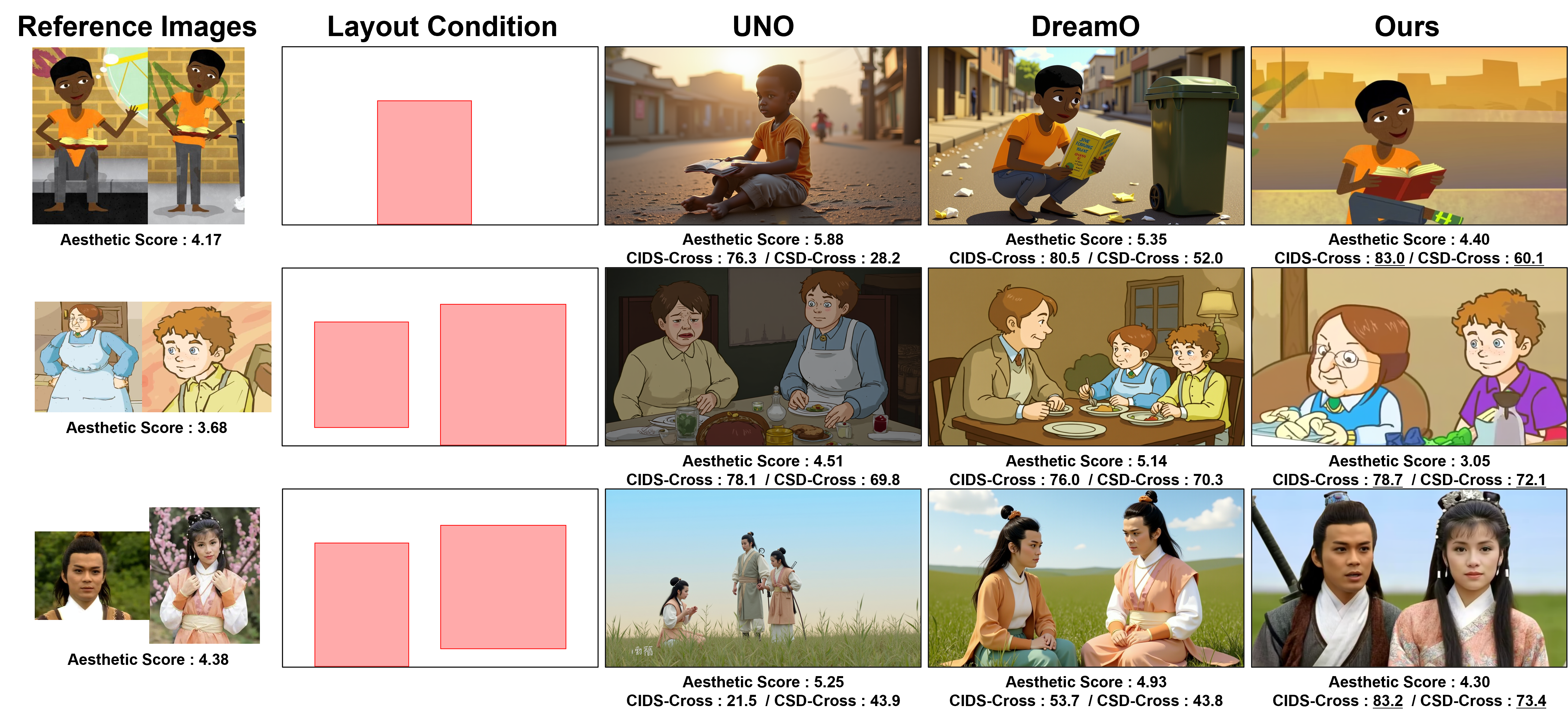}
    \vspace{-4mm}
    \caption{Image level qualitative comparison against other methods, along with their respective scores. Our method shows enhanced visual consistency with respect to the reference images, which is reflected in the similar aesthetic scores.}
    \label{fig:dreamingcomics_quality1}
    \vspace{-6mm}
\end{figure*}


\textbf{Layout Generation Model Training.} We finetune Qwen2.5-VL (7B)~\cite{yang2024qwen2technicalreport}, which has outperformed alternatives on comic layout understanding tasks~\cite{ViVoli_2026_ComicsPAP_ICDAR}, on 25K annotated comic layouts using supervised fine-tuning. Layouts are represented as dictionaries of normalized bounding boxes over fixed panel counts. We apply LoRA with rank 8, $\alpha=16$, dropout=0.05, and train using the AdamW optimizer (lr=5e-4).

\textbf{Dream-Illustrator Training.} We finetuned HunyuanVideo-I2V~\cite{Kong_2024_HunyuanVideo_arxiv} attached with the FramePack \cite{Zhang_2025_FramePack_arxiv} LoRA weights. To support multi-subject conditioning and reduce copy-paste artifacts, we remove the original image projection module. Training is performed using LoRA~\cite{Hu2021LoRALA} with rank 32, AdamW~\cite{Loshchilov2017AdamW} optimizer (lr=2e-4), batch size 8, and mixed precision on 2×NVIDIA H100 GPUs. The $\lambda_\text{mask}$ is set to 0.05. We train for 6K steps on single-subject samples, followed by 3K steps on multi-subject samples.

\subsection{Image Customization Evaluation} We evaluate DreamingComics using ViStoryBench~\cite{Zhuang_2025_VistoryBench_arxiv}, the first comprehensive story visualization benchmark. The benchmark primarily measures Character Similarity (CIDS) ~\cite{Jiankang_2019_ArcFace_CVPR, Kim_2022_Adaface_CVPR, Cao2017VGGFace2AD, Radford_2021_CLIP_ICML} and Style Similarity (CSD) ~\cite{Somepalli2024CSD} between reference and generated frames (cross) and between generated frames (self). The benchmark also defines Onstage Character Count Matching (OCCM), Inception Score (IS)~\cite{Salimans_2016_IS_NIPS}, and Aesthetic Score~\cite{AestheticPredictor}. To measure subject-layout fidelity, we add a layout precision metric for methods that support layout-based customization \cite{Wu_2025_DiffSensei_CVPR, Zhang_2025_Eligen_arxiv}. We also evaluate our method on DreamBench++ \cite{peng2024dreambench} for single-subject driven generation against other image customization methods. 
Further details on the evaluation procedure and metrics for quantitative evaluations are provided in \textbf{Supplementary}.




As shown in Table~\ref{tab:vistorybench}, DreamingComics achieves substantial improvements in both character and style preservation. It achieves the highest CIDS-cross score of 66.6, surpassing the next best method (DreamO~\cite{Mou_2025_DreamO_arxiv}, 51.5) by 29.2\%, and achieves a CSD-cross score of 53.6, highlighting our ability to \textbf{maintain both character identity and artistic style}. Our model also achieves an OCCM score of 86.7 and a layout precision score of 61.6, confirming correct layout-wise character generation. Although our Aesthetic Score and Inception Score are comparable to the copy-paste baseline, this reflects our deliberate preservation of stylized, non-photorealistic appearances—an area where video models, despite lower perceptual sharpness, excel in maintaining stylistic consistency (e.g., Fig.~\ref{fig:dreamingcomics_quality1}). We also present the results for DreamBench++ in Table~\ref{tab:dreambench_comparison}, which shows that our model achieves the highest DINO~\cite{Caron_2021_DINO_ICCV} and CLIP-I~\cite{Radford_2021_CLIP_ICML} scores, with CLIP-T~\cite{Radford_2021_CLIP_ICML} scores that are comparable to other methods.

\begin{table}[t]
\centering
\resizebox{0.8\columnwidth}{!}{
    \begin{tabular}{lccc}
    \toprule
    Method & DINO$\uparrow$ & CLIP-I$\uparrow$ & CLIP-T$\uparrow$ \\
    \midrule
    DreamBooth \cite{Ruiz2022DreamBoothFT} & 53.34 & 75.30 & 32.26 \\
    UNO~\cite{Wu_2025_UNO_ICCV} & 56.24 & 77.96 & 33.23 \\
    DreamO~\cite{Mou_2025_DreamO_arxiv} & 58.59 & 79.35 & 32.96 \\
    \midrule
    \textbf{Ours} & 60.50 & 79.50 & 32.59 \\
    \bottomrule
    \end{tabular}
}
\vspace{-2mm}
\caption{Quantitative results on DreamBench.}
\vspace{-6mm}
\label{tab:dreambench_comparison}
\end{table}

\begin{figure*}
\vspace{-3mm}
    \centering
    \includegraphics[width=\textwidth]{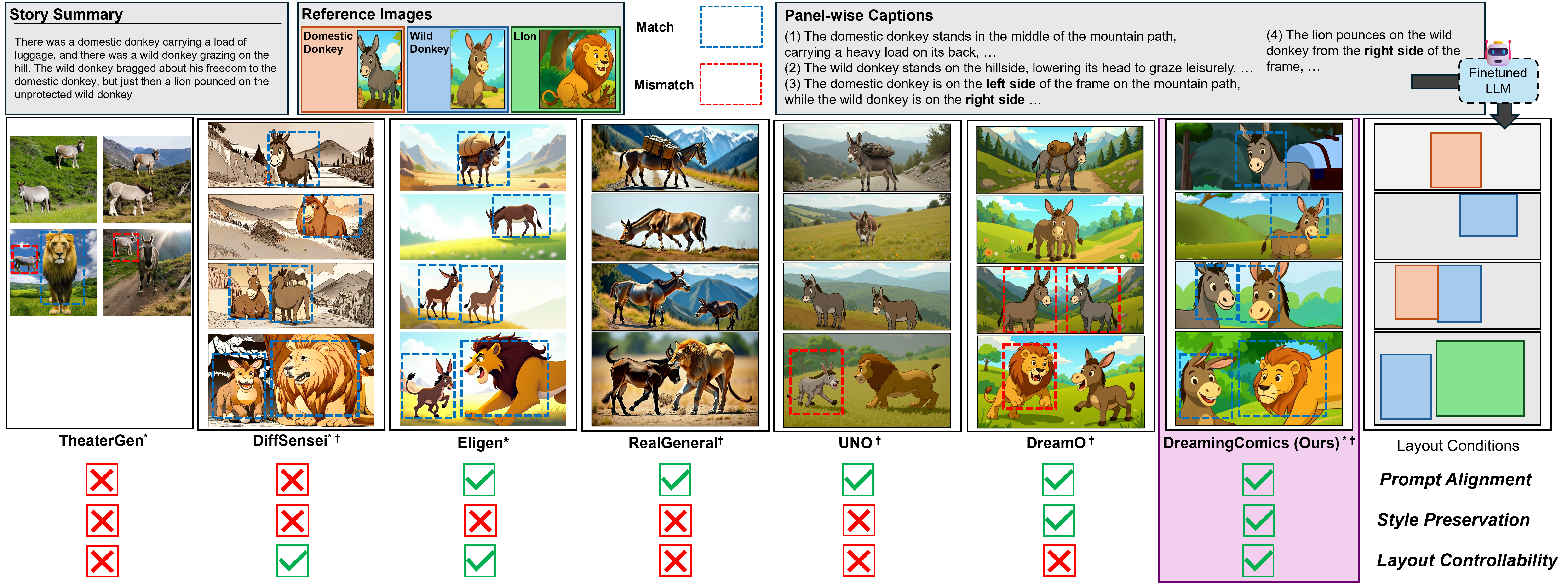}
    \vspace{-6mm}
    \caption{Story-level qualitative comparison of DreamingComics. Methods marked with ``$\ast$" use the layout condition generated from our layout generator. Methods marked with ``$\dagger$" use the reference images. Better viewed with zoom-in.}
    \label{fig:dreamingcomics_quality2}
    \vspace{-2mm}
\end{figure*}

Fig. \ref{fig:dreamingcomics_quality1} displays an image-level comparison with other methods \cite{Wu_2025_UNO_ICCV, Mou_2025_DreamO_arxiv}. Unlike other methods, our method utilizes layouts to designate each character's position, enhancing controllability while preventing superfluous or erroneous character creation (e.g., DreamO \cite{Mou_2025_DreamO_arxiv}, second and third rows). Also, the results from other methods produce higher aesthetic scores, even when their output style largely deviates from that of the reference's (e.g., first row). This is because aesthetic scoring tends to penalize images that do not look photorealistic rather than evaluating for diverse styles. However, \textit{our method can faithfully adhere to the given reference style, which is indicated by the similarity in aesthetic scores}. 

We also present a comic-level comparison in Fig. \ref{fig:dreamingcomics_quality2}, where we first generate layouts from the input panel-wise captions and use them along with the reference images and captions to synthesize a full-length comic-style story. Layout-based methods like DiffSensei \cite{Wu_2025_DiffSensei_CVPR} and Eligen \cite{Zhang_2025_Eligen_arxiv} fail to preserve the aesthetics of images, either because they do not allow image-level inputs or fail to preserve the input style. Image-based customization methods such as RealGeneral \cite{Lin_2025_RealGeneral_ICCV}, UNO \cite{Wu_2025_UNO_ICCV}, and DreamO \cite{Mou_2025_DreamO_arxiv} cannot utilize layout conditions, which creates confusion in placing characters. For instance, DreamO places the gray-colored ``Domestic Donkey" on the right side and the brown-colored ``Wild Donkey" on the left, which is not intended in its caption. In contrast, our method produces accurate and consistent story visualizations in terms of identity consistency, style preservation, prompt adherence, and layout composition.



\begin{table}
    \centering
    \resizebox{\columnwidth}{!}{\begin{tabular}{@{}lccccc@{}}
        \toprule
        \textbf{Method} & \textbf{Panel} & \textbf{Coverage}  & \textbf{Panel} & \textbf{Valid} & \textbf{Character} \\
          & \textbf{Count} & \textbf{Ratio} & \textbf{Ordering} & \textbf{Characters} & \textbf{Count} \\
        \midrule
        Ours & 100.0 & 79.14 & 99.00 & 100.0 & 86.30 \\
        GPT-4 \cite{Achiam2023GPT4TR} & 87.67 & 29.25 & 66.67 & 57.67 & 76.47 \\
        \bottomrule
    \end{tabular}
    }
    \vspace{-3mm}
    \caption{Evaluation results on our layout generator.}
    \vspace{-6mm}
    \label{tab:author_gpt}
\end{table}

\vspace{-1mm}
\subsection{Layout Generator Evaluation} We compare our layout generator with GPT-4 \cite{Achiam2023GPT4TR} to evaluate the validity of our method. We use a detailed prompt with in-context examples to guide GPT-4 to generate layouts as a list of bounding boxes, and evaluate the results according to the following categories:

\vspace{-1mm}
\begin{itemize}
    \item \textbf{Panel Count}: A correct number of panels should be present per page. 
    \item \textbf{Page coverage}: Panels should collectively cover the page without large empty regions.
    \item \textbf{Panel ordering}: Panels should follow the conventional comic order (top-to-bottom, right-to-left).
    \item \textbf{Valid Character}: The character bounding box should cover a reasonable area within the panel.
    \item \textbf{Character Count}: A correct number of character bounding boxes should be present per panel.
\end{itemize}
\vspace{-1mm}

We present the results in Table~\ref{tab:author_gpt}, which shows our model's improvements in understanding and generating comic-structured layouts.



\begin{table}[t]
\centering
\vspace{-3mm}
\setlength{\tabcolsep}{2pt}
    \resizebox{\linewidth}{!}{
        \begin{tabular}{cccc}
        \toprule
        \textbf{Regional RoPE} & \textbf{Masked Condition Loss} & \textbf{CIDS (Cross/Self)} & \textbf{CSD (Cross/Self)} \\
        \midrule
        \xmark  & \cmark & 38.7 / 50.3 & 41.0 / 52.9 \\
        \cmark & \xmark  & 56.6 / 61.2 & 50.5 / 56.3 \\
        \rowcolor{gray!10}
        \cmark & \cmark  & \textbf{58.5} / \textbf{63.0} & \textbf{52.9} / \textbf{62.4} \\
        \bottomrule
        \end{tabular}
    }
    \vspace{-2mm}
    \caption{Ablation study on the impact of RegionalRoPE and masked condition loss Our final model achieves the highest identity and style preservation performance without relying on regional attention masking.}
\label{tab:ablation_table}
\end{table}

\begin{table}[t]
\centering
\vspace{-3mm}
\setlength{\tabcolsep}{2pt}
    \resizebox{\linewidth}{!}{
        \begin{tabular}{lcc}
        \toprule
        \textbf{Name} & \textbf{CIDS (Cross/Self)} & \textbf{CSD (Cross/Self)} \\
        \midrule
        Timestamp $(t = 1)$ & 55.3 / 58.8 & 50.3 / 56.1 \\
        Timestamp $(t = 5)$ & 55.7 / 58.9 & 50.2 / 54.3 \\
        Timestamp $(t = 9)$ & 54.7 / 58.2 & 49.7 / 56.1 \\
        \rowcolor{gray!10}
        Ours $(t=3)$  & \textbf{58.5} / \textbf{63.0} & \textbf{52.9} / \textbf{62.4} \\
        \bottomrule
        \end{tabular}
    }
    \vspace{-3mm}
    \caption{Ablation study on the impact of choosing the right timestamp for the target image.}
    \vspace{-7mm}
\label{tab:ablation_table2}
\end{table}

\subsection{Ablation Study} We perform an ablation study to assess the contribution of our main components: RegionalRoPE and masked condition loss. As shown in Table~\ref{tab:ablation_table}, removing RegionalRoPE leads to the most significant performance drop, highlighting its role in spatial disentanglement. Excluding the masked condition loss also degrades layout fidelity. Overall, the combination of both components yields the best performance, validating our use of layout conditioning over rigid attention masking~\cite{chen2024trainingfree, Zhang_2025_Eligen_arxiv}. Additionally, Table~\ref{tab:ablation_table2} evaluates the effect of choosing the target timestamp $t$ for FramePack \cite{Zhang_2025_FramePack_arxiv}. We observe that $t = 3$ achieves the best trade-off between identity preservation and stylistic variation, as lower values produce rigid results, while higher values introduce undesirable drift. All models were trained for 6K steps on the single-subject samples and evaluated on ViStoryBench. We also include ablation studies on choosing $\lambda_\text{mask}$ values in \textbf{Supplementary}.

\subsection{User Study} To evaluate the subjective quality of our framework, we conducted a user study with 26 participants who were shown image pairs for four evaluation criteria: character identity, character style, story consistency, and layout plausibility. Each participant was presented with 20 comparison questions and was asked to select the preferred output. For the first 15 questions (three categories), the participants compared our results with those of DreamO \cite{Mou_2025_DreamO_arxiv} and UNO \cite{Wu_2025_UNO_ICCV}. The remaining questions compare our layout generator with GPT-4. The user study questionnaire is included in \textbf{Supplementary}. As shown in Table~\ref{tab:userstudy}, our work demonstrates a strong advantage in both image quality and layout control.

\begin{table}[t]
\vspace{-3mm}
    \centering
    \small
    \setlength{\tabcolsep}{4pt}
    \renewcommand{\arraystretch}{1.2}
    
    \resizebox{\columnwidth}{!}{ 
    \begin{tabular}{lcccc}
    \toprule
    \textbf{Method Preference (\%)} &
    \makecell{Character\\Identity} &
    \makecell{Character\\Style} &
    \makecell{Story\\Consistency} &
    \makecell{Layout\\Plausibility} \\
    \midrule
    Ours vs. Baseline & 80.0 & 83.8 & 86.2 & 69.2 \\
    \bottomrule
    \end{tabular}
    }
    \vspace{-2mm}
    \caption{
    User study results: percentage of participants who preferred our method over the baseline across four questions.
    }
    \vspace{-7mm}
\label{tab:userstudy}
\end{table}
\vspace{-2mm}

\section{Conclusion}
We present DreamingComics, a novel framework for subject- and layout-controllable story visualization. The problem is decomposed into layout generation and layout-based image customization. Our image customization module builds on a DiT video generation model, leveraging its spatiotemporal priors to ensure visual consistency across diverse artistic styles. For intuitive and semantically rich layout creation, we use a fine-tuned LLM-based layout generator. To enable layout-based control with multiple subjects, we introduce RegionalRoPE and a masked condition loss. Quantitative results and user studies demonstrate that DreamingComics produces faithful layouts and stylistically coherent visual narratives, establishing a strong foundation for future work in controllable story generation.



{
    \small
    \bibliographystyle{ieeenat_fullname}
    \bibliography{main}
}


\end{document}